\documentclass[10pt,twocolumn,letterpaper]{article}

\usepackage{iccv}
\usepackage[margin=4pt,font=small,labelfont=bf,labelsep=endash,tableposition=top]{caption}
\usepackage[colorlinks=true,linkcolor=blue,citecolor=blue]{hyperref}%

\usepackage{epsfig}
\usepackage{graphicx}
\usepackage{amsmath}
\usepackage{amssymb}
\usepackage{booktabs}
\usepackage{array,hhline}

\usepackage{multirow}
\usepackage{multicol}
\usepackage{subcaption}
\usepackage{enumitem,eucal}

\usepackage{comment}

\usepackage[symbol*]{footmisc}

\usepackage{color}
\usepackage{xspace}
\usepackage{textcomp}
\usepackage{cite}
\usepackage{mathtools}
\usepackage{bbm}
\usepackage{makecell, verbatim, sidecap}

\newcommand{\myparagraph}[1]{{\vspace{.5em} \noindent \bf #1}}

\def\eg{{\it e.g.}\xspace}
\def\ie{{\it i.e.}\xspace}
\def\Real{{\mathbb R}}

\def\x{{$\times$}}
\newcommand{\app}{\raise.17ex\hbox{$\scriptstyle\sim$}}

\def\Ours{{FCOS\textsubscript{PSS}}\xspace}
\def\RetinaNet{{ATSS}}
\def\ATSS{{ATSS}}
\def\CondInst{{CondInst}}

\def\OursATSS{{\ATSS\textsubscript{PSS}}\xspace}
\def\OursA{{\ATSS\textsubscript{PSS}}\xspace}
\def\OursCondInst{{\CondInst\textsubscript{PSS}}\xspace}

\def\Loss{ {\cal L} }

\def\centerness{{\rm ctr}}
\def\pss{{\rm pss}}

\def\OurTitle{{Object Detection Made Simpler by Eliminating Heuristic NMS}}

\iccvfinalcopy %

\begin{document}

\title{ \OurTitle }

\author{
Qiang Zhou$ ^{1\dag}$, ~~~~ Chaohui Yu$ ^{1\dag}$, ~~~~ Chunhua Shen$ ^{2\dag}$, ~~ ~~ Zhibin Wang$ ^1$, ~~~~ Hao Li$ ^1$\\[0.2cm]
$^1 $ Alibaba Group  ~ ~ ~ ~ ~ ~ ~ ~ $^2 $ Monash University; The University of Adelaide
}

\makeatletter
\let\@oldmaketitle\@maketitle%
\renewcommand{\@maketitle}{\@oldmaketitle%
 \centering
    \includegraphics[width=.8851\linewidth]{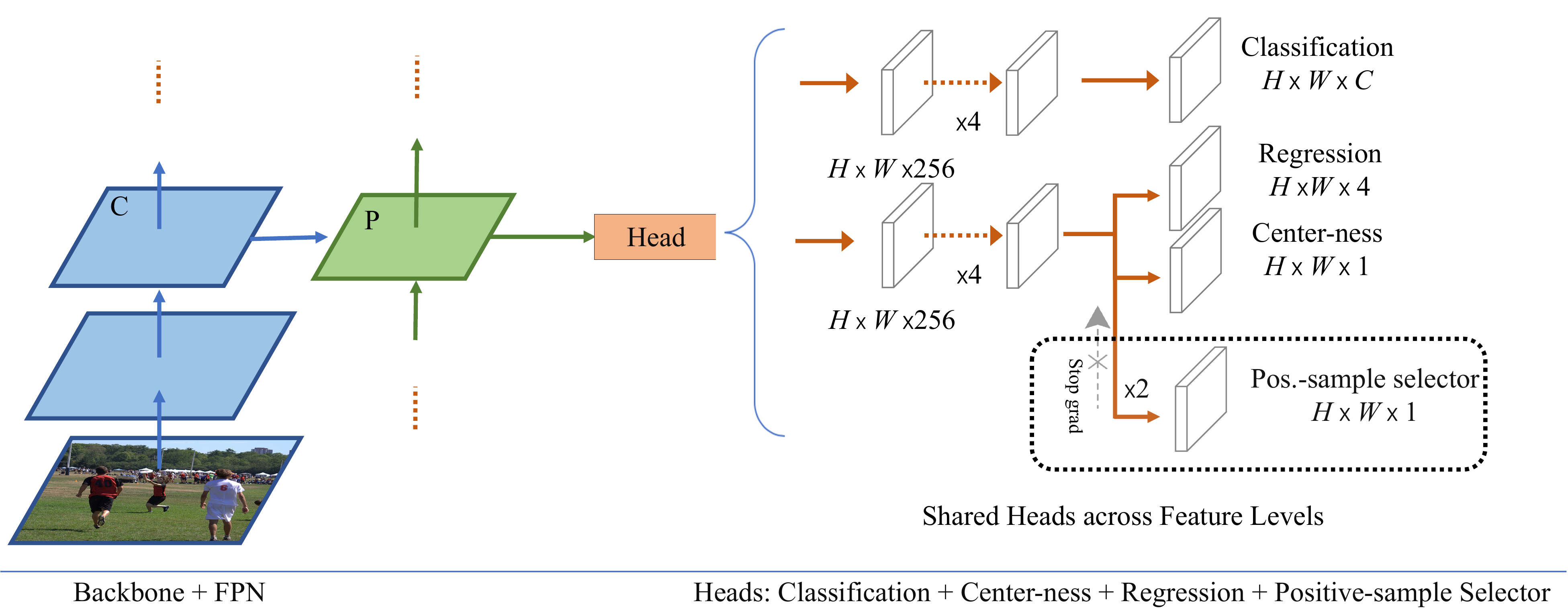}
    \captionof{figure}{
    \textbf{The proposed detector \Ours, which is NMS free and end-to-end trainable}.  
     Compared to the original FCOS detector, the only  modification  to the
     network
     is the
     introduction of the `positive sample selector (PSS)' as shown in the dashed box. Because the PSS head consists of only two compact conv.\ layers, the computation overhead is negligible  (\app 8\%).
     Here the `Stop-grad' operation plays an important role in training (see details in the text \S\ref{subsec:stopgrad}).
    }
    \label{fig:model}
  \bigskip}
\makeatother

\maketitle

\def\thefootnote{$\dag$}\footnotetext{Equal contributions.}\def\thefootnote{\arabic{footnote}}

\begin{abstract}

     We show a simple NMS-free, end-to-end object detection framework, of which the network is a minimal modification to a one-stage object detector such as the FCOS detection model \cite{tian2019fcos}.  
     We attain {\rm  on par} or even improved detection accuracy compared with the original one-stage detector. It performs detection at almost the same inference speed, while being even simpler in that now
     the post-processing NMS (non-maximum suppression) is eliminated during inference.
     If the network is capable of identifying only one positive sample for prediction for each ground-truth object instance in an image, then NMS would become unnecessary. 
     This is made possible by attaching a compact PSS head for automatic selection of the single positive sample for each instance (see Fig.~\ref{fig:model}). 
     As the learning objective involves both one-to-many and one-to-one label assignments, there is a conflict in the labels of some training examples, making the learning  challenging. 
     We show that by employing a stop-gradient operation, we can successfully 
     tackle this issue and train the detector.

     On the COCO dataset, our simple design achieves 
     superior 
     performance compared to both the  FCOS baseline detector 
     with NMS post-processing and the recent 
     end-to-end NMS-free  detectors.
     Our extensive ablation studies justify the rationale of the design choices. 

    {   \def\UrlFont{\sf}
    \def\UrlFont{\rm\small\ttfamily}
     Code is available at: \url{https://git.io/PSS} 
     }

\end{abstract}

\section{Introduction}

Object detection is a fundamental task in computer vision and has 
been progressed 
dramatically  in the past few years using deep learning.  It aims at predicting a set of bounding boxes and the corresponding categories in an image. 
Modern object detection methods fall into two categories: two-stage detectors, exemplified by Faster R-CNN \cite{faster-rcnn}, and one-stage detectors such as 
YOLO \cite{yolo}, SSD \cite{SSD}, RetinaNet \cite{LinGGHD17_retinanet}. 
Recently one-stage detectors become more and more popular due to its simplicity and high performance. Since anchor-free detectors such as FCOS \cite{tian2019fcos} and 
FoveaBox \cite{kong2019foveabox} were introduced,  the community tend to recognize that anchor boxes may not be an indispensable design choice for object detection, thus leaving NMS (non maximum suppression) the only heuristic post-processing in the entire pipeline.   

The NMS operation has almost always been used  in mainstream object detectors in the literature. The necessity  of NMS in detection is because one ground-truth
object always has multiple positive samples during the course of training. For instance, in \cite{tian2019fcos,kong2019foveabox}, all the locations on the CNN feature maps within the center region of an object are assigned  positive labels.  As a result, multiple network outputs correspond to one target object.  The consequence is that for inference, a mechanism (namely, NMS) is needed to choose the best positive sample among all the positive boxes.

Very recently,  some methods \cite{CarionMSUKZ20_detr,zhu2020deformable} formulate  detection  into  a  set-to-set  prediction  problem  and  leverage the Hungarian matching algorithm  to tackle the issue of finding one positive sample for each ground-truth box. With the Hungarian matching algorithm, for a ground-truth object,  the models adaptively choose the network outputs with  minimal  loss  as  the  positive  sample  for  the  object, while  other  samples  are  considered  negative. Thus NMS can be removed and the detector becomes end-to-end trainable. 
Wang \etal \cite{wang2020end_defcn} design a fully convolutional network (FCN) without using Transformers and achieve end-to-end NMS-free object detection. 
Our work here follows this avenue by designing even simpler FCN detectors with stronger detection accuracy.

Concretely, in this work we want to design  \textit{a  simple high-performance fully convolutional network for object detection, which is NMS free and fully end-to-end trainable. }  
We instantiate such a design on top of the FCOS detector \cite{tian2019fcos}
with minimal modification to the  network itself,
as shown in Fig.~\ref{fig:model}. Indeed, the only modification to FCOS is that \textit{i}.)
a compact ``\textit{positive sample selector} (PSS)'' head is introduced, in order to select the optimal positive sample for each object instance; \textit{ii}.) the learning objective is re-designed to successfully train the detector.

For training the network, we keep the original FCOS classification loss, which is important as it provides rich supervision that encodes desirable invariance encoding.
As discussed in \S\ref{subsec:conflict}, the  label discrepancy  
between one-to-many and one-to-one label assignments causes the network training 
challenging.  Here,  we propose a simple asymmetric optimization scheme:
we introduce the {\tt stop-grad}  operation (see Fig.~\ref{fig:model})  to stop the gradient relevant to the attached PSS head passing to the original FCOS network parameters.

 We empirically show the effectiveness of the {\tt stop-grad} 
   operation (Fig.~\ref{fig:stopgrad}). This {\tt stop-grad} can be particularly useful in the following case. The network contains two sub-networks {\tt A} and {\tt B}.\footnote{In this work, {\tt A} is the FCOS network, and { \tt B}
   is the PSS head.}
   The optimization of {\tt B} relies on the convergence of {\tt A} while the optimization of {\tt A} does not rely \textit{much} on {\tt B}, thus being asymmetric.

 Our method does not rely attention  mechanisms such as Transformers,
 or %
 cross-level feature aggregation such as the 3D max filtering 
 \cite{wang2020end_defcn}. Our proposed detector, termed \Ours,  enjoys the following advantages. 
\begin{itemize}
\itemsep -0.1cm

\item 
    Detection  is  now  made  even  simpler  by  eliminating NMS. With the simplicity inherited from FCOS, \Ours is fully compatible with other FCN-solvable tasks.

\item 
We show that NMS can be eliminated by introducing a single compact PSS head, with negligible computation overhead compared to the vanilla FCOS.

\item 
    The proposed PSS is flexible in that, in essence the PSS head
    serves as \textit{learnable NMS}. Because by design we keep the vanilla FCOS heads working as well as the original detector, \Ours\ offers flexibility in  terms of the NMS being used. For example,  once trained,
    one may choose to discard the PSS head and use \Ours\ as the standard FCOS.

\item 
  We report \textit{on par} or even improved detection results on the COCO dataset compared with the standard FCOS \cite{tian2019fcos} and ATSS
  \cite{zhang2020bridging_atss} detectors, as well as  
  recent NMS-free detection methods.
 \item 
  The  proposed  PSS head  is  also  applicable  to  other anchor-box based detectors such as RetinaNet. We achieve promising results by attaching PSS to  the 
   modified RetinaNet detector, which uses one square anchor box per location and employs adaptive training sample selection as for improved detection accuracy as in \cite{zhang2020bridging_atss}\footnote{Hereafter we refer to this model  as \ATSS; and 
   the corresponding version with a PSS head is 
   \OursA.}.
   
   \item 
  
   The same idea may  be applied to some other instance recognition tasks. For example, 
   as we shown in \S\ref{exp:inst}, 
   we can eliminate NMS  from    instance  segmentation  frameworks, \eg,   \cite{wang2020solo,SoloV22020}.
 We wish that the work presented here can benefit those works built upon the FCOS detector 
including instance 
segmentation \cite{chen2020blendmask, tian2020conditional, zhang2020MEInst,xie2020polarmask},  keypoint detection \cite{DirectPose},
text spotting \cite{Liu2020ABCNet}, and tracking \cite{SiamCAR}.   
\end{itemize}

\section{Related Work}

Object detection has been extensively studied in computer vision as it enables a wide range of downstream applications. 
Traditional methods use hand-crafted features (\eg, HoG, SIFT) to solve detection 
as classification 
on a set of candidate bounding boxes.
With the development of deep neural networks,
modern object detection methods can be divided into three categories: two-stage detectors, one-stage detectors, and recent end-to-end detectors.

\myparagraph{Two-stage Object Detector}
One line of research focuses on two-stage object detectors such as Faster R-CNN \cite{faster-rcnn}, which first generate region proposals, and then refine the detection for each proposal. 
Mask R-CNN \cite{he2017mask} adds a mask prediction branch on top of Faster R-CNN, which can be used to solve a few instance-level recognition tasks, including instance segmentation and pose estimation.

While two-stage detectors still find many applications, the community have been shifting the research focus to one-stage detectors due to its much simpler and cleaner design, with strong performance.  

\myparagraph{One-stage Object Detector}
The second line of research develops efficient single-stage object detectors \cite{SSD,yolo,tian2019fcos,kong2019foveabox}, which directly make dense predictions based on extracted feature maps and dense anchors or points.
They are based essentially sliding windows.
Anchor-based detectors, \eg, YOLO~\cite{yolo} and SSD~\cite{SSD} use a set of pre-defined anchor boxes to predict object category and anchor box offsets.
Note that anchors were first proposed in the RPN module of Faster R-CNN to generate proposals. 

In one-stage object detection, negative samples are inevitably many more than positive samples, leading to data imbalance training. 
Techniques such as hard negative mining and the focal loss function \cite{LinGGHD17_retinanet} were thus proffered to alleviate the imbalance training issue.

Recently, efforts have been spent on designing \textit{anchor-free} detectors \cite{tian2019fcos,kong2019foveabox}. 
FCOS \cite{tian2019fcos} and Foveabox \cite{kong2019foveabox} use the center region of targets as positive samples.
In addition, FCOS introduces the so-called center-ness score to make NMS more accurate.
Authors of \cite{zhang2020bridging_atss} propose an adaptive training sample selection (ATSS) scheme to automatically define positive and negative training samples, albeit still using a heuristically designed rule. PAA \cite{kim2020probabilistic_paa} designs a probabilistic anchor assignment strategy, leading to easier training compared to those heuristic IoU hard label assignment strategies.
Besides improving the assignment strategy of FCOS \cite{zhang2020bridging_atss,kim2020probabilistic_paa}, efforts were also spent on 
the detection features \cite{QiuMLLS20}, loss functions \cite{li2020generalized}   to 
further boost the anchor-free detector's
performance.

Nevertheless, both the one-stage and two-stage object detectors require a  post-processing procedure, namely, non-maximum suppression (NMS), to merge the duplicate detections. That is, most state-of-the-art detectors are not end-to-end trainable.

\myparagraph{End-to-end Object Detector}
Recently, a few works propose end-to-end frameworks for object detection by removing NMS from the pipeline. 
One of the pioneering works may be attributed to \cite{Stewart2016end2end}.
There, object detection was formulated as a sequence-to-sequence learning task and LSTM-RNN was used to implement the idea. 
DETR~\cite{CarionMSUKZ20_detr} introduces the Transformer-based attention mechanism to object detection.
Essentially the sequence-to-sequence learning task in \cite{Stewart2016end2end} 
was now solved in parallel by 
self-attention based Transformer rather than RNN.
Deformable DETR \cite{zhu2020deformable} accelerates 
the training convergence of DETR by 
proposing to only perform attention to a small set of key sampling points.

Very recently, DeFCN  \cite{wang2020end_defcn}  adopts a one-to-one matching strategy to enable end-to-end object detection based on a fully convolutional network with competitive performance. 
Significantly, probably for the first time, DeFCN \cite{wang2020end_defcn} demonstrates that it is possible to remove NMS from a detector without resorting to sequence-to-sequence (or set-to-set) learning that relies on LSTM-RNN or self-attention mechanisms.  
The work in \cite{OneNet} shares similarities with 
DeFCN \cite{wang2020end_defcn} in the one-to-one label assignment and 
using auxiliary heads to help training. The performance reported in \cite{OneNet} is inferior to that of DeFCN \cite{wang2020end_defcn}.

We have drawn inspiration from \cite{wang2020end_defcn} in terms of designing the one-to-one label assignment strategy, as in Equ.~\eqref{eq:quality}. Clearly, this 
one-to-one label assignment has a direct impact on the final NMS-free detection accuracy. 
One of the main differences is that we use a simple binary classification head to 
enable selection of one positive sample for each instance,
while DeFCN designs 3D max filtering for this purpose, which aggregates multi-scale features to suppress duplicated detections. 
Second, 
As mentioned earlier,
\textit{we have  aimed to keep the original FCOS detector intact  as much as possible. 
}
Next, we present the proposed \Ours.

\section{Our Method}

\begin{table*}[h!]
\small
\centering 
    \begin{tabular}{r | r |c|c|c|c}
    \toprule
    Backbone                    & Model           & mAP (\%)   & mAR (\%)  & Network forward (ms) & Post process.\  (ms) \\ 
    \midrule
    \multirow{5}{*}{R50}  & FCOS~\cite{tian2019fcos}            & 42.0 & 60.2 & 38.76 & 2.91   \\ %
                                & \RetinaNet~\cite{zhang2020bridging_atss}            & 42.8 & 61.4 & 38.31 & 7.56    \\  
                                & DeFCN~\cite{wang2020end_defcn}           & 41.5 & 61.4 &   &  \\ 
                                & 
                                  \textbf{\Ours} (Ours) & 42.3 & 61.6 &  42.37 & 1.49   \\ 
                                &  
                                 {\bf \OursATSS} (Ours)
                                   & 42.6 & 62.1 &  42.19 & 3.89   \\ \midrule
    \multirow{4}{*}{R101} & FCOS  \cite{tian2019fcos}          &  43.5    &   61.3   & 51.55  & 3.40 \\  
                                & \RetinaNet \cite{zhang2020bridging_atss}            &  44.2    &  61.9    & 51.81  & 8.80  \\  
                                & 
                                  \textbf{\Ours} (Ours) &  44.1    &   62.7   &  54.95 & 2.52  \\  
                                & 
                                {\bf \OursATSS} %
                                (Ours) &   44.2   &    63.2  & 55.65  & 4.32 \\ \midrule
    X-101-DCN & %
                {\bf \OursATSS} (Ours) & 47.5 & 65.1 & 82.64 & 4.31 \\ \midrule
    R2N-101-DCN & %
                    {\bf \OursATSS} (Ours) & 48.5 & 66.4 & 81.30 & 4.17 \\ \bottomrule
    \end{tabular}
\caption{\textbf{Performance comparison 
between our proposed NMS-free detectors and 
various one-stage detection methods}
on the COCO val.\  set. All the models are trained with the `3\x' schedule and multi-scale data augmentation (but with single-scale testing).
Here `\RetinaNet'  is the modified RetinaNet detector using one square anchor box per location, employing ATSS (adaptive training sample selection) as in \cite{zhang2020bridging_atss}. Inference time  measures network forward computation and post-processing time (ranking detected boxes to compute mAP for our methods and NMS for others) on a single V100 GPU.
Backbones: `R': ResNet~\cite{he2016deep}. `X': ResNeXt (32\x4d-101) \cite{XieGDTH17}. `DCN': Deformable Convolution Network~\cite{zhu2019deformable}. `R2N': Res2Net~\cite{Shanghua2019}.}
\label{tab:main} 
\end{table*}

The overall network structure of \Ours 
is very straightforward as shown in Fig.~\ref{fig:model}.
The only modification is the PSS head which has two extra conv.\ layers,
and all the other parts are the same as that of FCOS \cite{tian2019fcos}.
We start by presenting the overall training objectives.

\subsection{Overall Training Objective}

Our overall training targets can be formulated as follows.
\begin{equation}
\Loss =   \Loss_{fcos} + \lambda_1  \cdot  \Loss_{pss} +  \lambda_2  \cdot \Loss_{rank}. 
\label{eq:overall_loss} 
\end{equation}
Here 
$ \lambda_1$, $ \lambda_2$ are balancing coefficients. 
$ \Loss_{fcos}$ contains the loss terms that are exactly same as in the original FCOS \cite{tian2019fcos}, namely, $C$-way  classification using the focal loss ($ C=80 $ for the COCO dataset), bounding box regression with the GIoU loss,
and the center-ness loss.  

We have set $ \lambda_2 $ to $ 0.25$ in all the experiments as the ranking loss is not essential.
As reported in Table~\ref{tbl:rankingloss}, the ranking loss improves the final detection accuracy slightly (0.2\app 0.3 points in mAP).  We include it here as it does not introduce much training complexity.

\subsubsection{PSS Loss $\Loss_{pss} $}

$ \Loss_{pss} $ is the key to the success of  our framework, which is a 
classification loss associated to the positive sample selector. 
Recall that our goal is to select one and only one positive sample for each instance in an image. Here the newly added PSS head is expected to achieve this goal. 
We defer the details of one-to-one positive label assignment to \S\ref{subsec:one2one} and let us assume that the one optimal ground-truth positive label for each instance is available.     
As shown in Fig.~\ref{fig:model}, the output of the PSS head is  a map of $ \Real^{H \times W \times 1} $. Let us denote $ \sigma (\pss ) $ a single point of this map.
Then the learning target for $ \sigma (\pss ) $ is $1$ if it corresponds to the 
positive sample of an instance; otherwise negative labels. Thus, the simplest approach would be to train a binary classifier. Here, in order to take advantage of the  $C$-way classifier in FCOS, we again formulate the loss term  as $ C $-way classification.

Specifically,  the loss term calculates the focal loss between the multiplied 
score of:\footnote{
We have included the center-ness score
$\sigma (  \centerness  )$
here to make it compatible with the original FCOS. As shown in FCOS \cite{tian2019fcos}, center-ness 
may
slightly improve the result.
} 
\[
\sigma (\pss ) \cdot \sigma (s) \cdot \sigma (  \centerness  )
\]
against the ground-truth labels of $ C $ categories, 
with $ \sigma(s) $ being the score map of the FCOS classifier's output.
Note that, the  difference between this classifier and the vanilla FCOS' classifier is that here we now have only one positive sample for each object instance in an image. After learning, ideally, $ \sigma( \pss ) $ is able to activate one and only one positive sample for an object instance.

\subsubsection{Ranking Loss $ \Loss_{ rank } $}

Our pilot experiment show that %
including a 
ranking loss term in %
the objective function 
can help improve the performance of our NMS-free detectors. Specifically, we add the ranking loss of Equ.~\eqref{eq:rank} for each training image:
\begin{equation}
\label{eq:rank}
    \Loss_{rank} = \frac{1}{n_{-}  n_{+}}\sum_{i_{-}}^{n_{-}}\sum_{i_{+}}^{n_{_+}}\text{max}(0, \gamma - \hat{P}_{i_{+}}(c_{i_{+}}) + \hat{P}_{i_{-}}(c_{i_{-}}))
\end{equation}
Here, $\gamma$ is the hyper-parameter representing the margin between positive and negative anchors. In our experiments, we set $\gamma = 0.5$ by default.
 {
    $n_{-}$ and $n_{+}$ denote 
    the number of negative and positive samples respectively. $\hat{P}_{i_{+}}(c_{i_{+}})$ denotes the classification score of positive sample $i_{+}$ belonging to category $c_{i_{+}}$, $\hat{P}_{i_{-}}(c_{i_{-}})$ denotes the classification score of negative sample $i_{-}$ belonging to category $c_{i_{-}}$. In our experiments, we choose the top $n_{-}$ scored $\hat{P}_{i_{-}}(c_{i_{-}})$ from all negative samples and we set $n_{-} = 100$ %
    for all the experiments.
    }

\subsection{One-to-many Label Assignment} 

    One-to-many label assignment---each object instance in a training image is assigned with  multiple ground-truth bounding boxes---%
is the widely-used approach to tackle the task of object 
detection.  This is the most intuitive formulation as naturally there is ambiguity in labelling the ground-truth bounding boxes for detection: if one shifts a labelled box by a few pixels, the resulting box still represents the same instance, thus being counted as a positive training sample too.
    That is also the reason why for the one-to-one label assignment, a static 
    rule (a rule that is irrelevant  to the prediction quality during training)   
    is less likely to produce satisfactory results because it is difficult to find well-defined annotations. 

    The advantage of having multiple boxes for one instance is that such a rich representation improves learning 
    of 
    a strong classifier that encodes invariances of  aspect ratios, translations, \etc.

    The consequence is multiple detection boxes around one single true instance.
Thus, NMS becomes indispensable, in order to clean up the raw outputs of such detectors.    
    Albeit  relying on heuristic NMS post-processing, we believe that 
    this one-to-many label assignment has its critical importance in that \textit{i.})
    richer training data help learning of some helpful invariances that are 
    proven beneficial;
    \textit{ii.})
    it is also consistent with the \textit{de facto} practice of 
    current 
    \textit{data augmentation} 
    in almost all deep learning techniques.
    Thus, we believe that it is important to keep the FCOS training target related to the one-to-many label assignment
    as an essential component of our new detector.

\subsection{One-to-one Label Assignment}
\label{subsec:one2one}

When performing one-to-one label assignment, we need to select the best matching anchor $i$ (here anchor represents an anchor box or an anchor point in different detectors) for each ground-truth instance $j$ (with $  c_j$ and $ b_j  $ being the ground-truth category label and bounding box coordinates).

As pointed out in \cite{wang2020end_defcn},  the best matching should include classification matching and location matching. That is, the classification 
quality and localization %
quality of the current detection network---during the course of training---should both %
play a role
in the matching score $Q_{i, j}$. 
We define the score as in
Equ.~\eqref{eq:quality}:
\begin{equation}
\label{eq:quality}
\begin{split}
Q_{i, j} = &\underbrace{\mathbbm{1} [i \in \Omega_j] }_{\rm positiveness~prior} \cdot  \underbrace{ 
\Bigl[ \hat{P}_{i}(c_j) \Bigr] ^{1-\alpha}}_{\rm classification} \cdot \underbrace{
\Bigl[
{\rm IoU}
(
\hat{b}_{i}, b_j)
\Bigr] 
^{\alpha}}_{\rm localization}.
\end{split}
\end{equation}
where
\begin{equation}
\label{eq:final_cls}
\hat{P}_i (c_j ) = \sigma( \pss_i)  \cdot \sigma(s_i) \cdot \sigma(\centerness_i).
\end{equation}

Here $s_i$ and $\centerness_i$ denote the classification score and center-ness prediction of anchor $i$.
Moreover, 
$ \pss_i $ denotes the binary mask prediction scores,
which is the output of the positive sample selector (PSS).
$\hat{P}_i( c_j)$ denotes the classification score of anchor $i$ belonging to category $ c_j $. 
Note that,  $ \sigma(\cdot) \in [0, 1] $ is the {\tt sigmoid} function that normalizes a score into a probability.  
In Equ.~\eqref{eq:final_cls}, we have assumed that the three probabilities are 
independent such that their  product forms $ \hat P_i ( c_j ) $. 
This may not hold strictly.

Now $Q_{i,j}$ represents the matching score between anchor $i$ and ground-truth instance $j$. 
$c_j$ is the ground-truth category label of  instance $j$.
$\hat{P}_i(c_j)$ denotes the prediction score of anchor $i$ corresponding to category label $c_j$.
$\hat{b}_i$ denotes predicted bounding box coordinates of anchor $i$; and 
$b_j$ denotes the ground-truth bounding box coordinates of  instance $j$.
The hyper-parameter $\alpha \in [0, 1]$ is used to adjust the ratio between classification and localization.

Given a ground truth instance $j$, not all anchors are suitable for assigning as positive samples, especially those anchors outside the box region of ground truth instance $j$. Here
$\Omega_j$  represents the set of the candidate positive anchors for instance $j$. 
FCOS~\cite{tian2019fcos}  restricts $\Omega_j$ to include only anchor points located in the center region of $b_j$ and RetinaNet~\cite{LinGGHD17_retinanet} %
restricts the IoU between $b_j$ and anchors in $\Omega_j$. The design for $\Omega_j$ would ultimately affect the performance of the model. For example, ATSS~\cite{zhang2020bridging_atss} and PAA~\cite{kim2020probabilistic_paa} further improve the performance of the FCOS~\cite{tian2019fcos} model by improving the design of $\Omega_j$ without changing the model structure. 

As shown in Equ.~\eqref{eq:quality}, in our case $\Omega_j$ is simply the positive samples used by the original detectors.
For example, \Ours uses anchor points in the center region of instance $j$ as $\Omega_j$ same as in FCOS; and $ \Omega_j $ of 
\OursA 
uses the ATSS sampling strategy in \cite{zhang2020bridging_atss}.

Finally, each ground-truth instance $ j $ in an image is assigned to one label by solving the bipartite graph matching using the Hungarian algorithm 
as in \cite{CarionMSUKZ20_detr,wang2020end_defcn}, 
by maximizing the quantity  
$ \sum_j Q_{ i , j }  $ by finding the optimal anchor index $ i $ for each instance $ j$.
We have observed similar performance if using the simple top-one  selection to replace the Hungarian matching.

\subsection{Conflict in the Two Classification Loss Terms}
\label{subsec:conflict}

Recall that in the overall objective Equ.~\eqref{eq:overall_loss}, we 
minimize two correlated classification objective terms. The first one is the 
vanilla FCOS classification (one-to-many) in $ \Loss_{fcos}$. Assume that a particular object
instance is assigned with $ k $ positive samples (anchor boxes or points).   
The second classification is in the PSS term $ \Loss_{pss} $. 
The main responsibility of $ \Loss_{pss} $ is to distinguish one and only one positive sample
(from the $ k $ positive samples) from the rest. 
Therefore,  when training the PSS classifier, $ k-1$ of the 
$ k $ positive samples for $ \Loss_{fcos} $ are given ground-truth labels of being negative. 
This potentially makes the fitting more challenging as 
labels are inconsistent for these two terms.

In other words, a few samples are assigned as positive samples and negative samples at the same time when training the model. This conflict may adversely impact the final model performance.
In this work, we propose a simple and effective asymmetric optimization scheme. Specifically, we stop the gradient relevant to   
the attached PSS head (the dashed box in Fig.~\ref{fig:model}) 
passing to the original FCOS network parameters (network excluding the dash box).  
This is marked as ``{\tt stop-grad}'' in Fig.~\ref{fig:model}.
Thus, the new PSS head would have minimal impact\footnote{But not zero 
impact 
because 
$ \Loss_{ pss } $ is coupled with the FCOS classifier in $ \Loss_{fcos}$.
}
on the training of the 
original FCOS detector.

%
%
%

\def\btheta{{\boldsymbol \theta }}
\def\argmin{{\rm argmin}}

\subsection{Stop Gradient}
\label{subsec:stopgrad}

Mathematically the stop-gradient operations sets a part of the network to be constant
during training
\cite{stopgrad}.
In our case, when SGD updating the vanilla FCOS parameters, 
the PSS head is set to be constant, thus zero gradients from PSS go backward to the remaining part of the network.

Let $ \btheta = \{ \btheta_{fcos}, \btheta_{pss} \} $ be all the network variables
for optimization, which splits into two parts. 
Essentially what stop-gradient does is similar to alternating optimization for these two sets of variables. 
We want to solve 
\[
\min_{  \btheta_{fcos}, \btheta_{pss} } \Loss (\btheta_{fcos}, \btheta_{pss}).
\]
We can solve two sub-problems alternatively ($t$ indexes iterations): 
\begin{equation}
 \btheta_{fcos}  ^ t  \leftarrow \argmin_{   \btheta_{fcos}  }
    \Loss (\btheta_{fcos}, \btheta_{pss}^{t-1} );
    \label{eq:alt1}
\end{equation}
and, 
\begin{equation}
 \btheta_{pss}  ^ t  \leftarrow \argmin_{   \btheta_{pss}  }
    \Loss (\btheta_{fcos}^{t}, \btheta_{pss} ).
    \label{eq:alt2}
\end{equation}
When solving for Equ.~\eqref{eq:alt1}, gradients w.r.t.\  $ \btheta_{pss} $ 
are zeros.
Note that SGD with stop-gradient as we do here is only approximately similar to the above alternating optimization, as we do not solve each of the two sub-problems to convergence. 

We may solve the two sub-problems for only one alternation. That is,
initializing  $ \btheta_{pss} = \boldsymbol 0 $ and solving Equ.~\eqref{eq:alt1} until convergence, then solving  Equ.~\eqref{eq:alt2} until convergence.  
This is equivalent to training the original FCOS until convergence and freezing FCOS, and then training the PSS head only until convergence. 
Our experiment shows that this leads to slightly %
inferior 
detection performance, but
with significantly longer training computation time.
See \S\ref{subsec: stopvstwostep} for details.

We empirically show that the use of stop gradient results in
consistently
improved 
model accuracy, referring to Fig.~\ref{fig:stopgrad}.

\section{Experiments}
In this section, we test our proposed methods on the large-scale dataset COCO~\cite{lin2014microsoft}.

\subsection{Implementation Detail}
We implement all our models 
using 
the MMDetection toolbox~\cite{mmdetection}.
For a fair comparison, we use the NMS-based detection methods~\cite{tian2019fcos,zhang2020bridging_atss} as the baseline  detectors and 
attach
our positive sample selector (PSS) to the work for eliminating 
the post-processing NMS. All the ablation %
comparisons 
are based on the ResNet50~\cite{he2016deep} backbone with FPN~\cite{lin2017feature}, and the feature weights are initialized by the pretrained ImageNet  model.
Unless otherwise specified, we train all the models with the `3\x' training schedule (36 epochs). %
Specifically, 
we train the models using  SGD %
on 8 Tesla-V100 GPUs, with an initial learning rate of 0.01, a momentum of 0.9, a weight decay of $10^{-4}$, a mini-batch size of 16. 
The learning rate decays by a factor of 10 %
    at the 24$^{\rm th}$ and 33$^{\rm th}$ epoch respectively.

\begin{table}[t!]
\begin{center}
\footnotesize 
    \begin{tabular}{ r  | c|c|c}
    \toprule
    \multirow{3}{*}{Model}    & \multirow{3}{*}{\tt Stop-grad}       & \multicolumn{2}{c}{mAP (\%)}  \\ \cline{3-4} 
                              &                                  & \begin{tabular}[c]{@{}c@{}}end-to-end pred.\\ 
                               (w/o NMS)\end{tabular} & \begin{tabular}[c]{@{}c@{}}
                              one-to-many %
                              pred.\\ (w/ NMS)\end{tabular} \\ \midrule
    \multirow{2}{*}{  \Ours\  } &                       &    41.5          &    41.2                 \\ 
                                & \checkmark            &    42.3          &    42.2                 \\ \midrule
    \multirow{2}{*}{  %
                      \OursATSS } &                    & 41.6             &  41.2                  \\  
                                   & \checkmark         & 42.6             & 42.3                    \\ \bottomrule
    \end{tabular}
\end{center}
\caption{Comparison of detection accuracy for %
training 
with and without `stop-gradient' on the COCO val.\  set. All models are trained with the `3\x' schedule and multi-scale augmentation. `end-to-end' prediction is the result of using PSS. `one-to-many' is the result by discarding the PSS head.
}
\label{tbl:ablat_detach}
\end{table}

\subsection{Ablation Studies}
Our main results are reported in Table \ref{tab:main}. 
We first present ablation studies to empirically justify the design choices.

\def\widvis{0.13794}

\begin{figure*}[t!]
\centering 
\includegraphics[width=\widvis\linewidth]{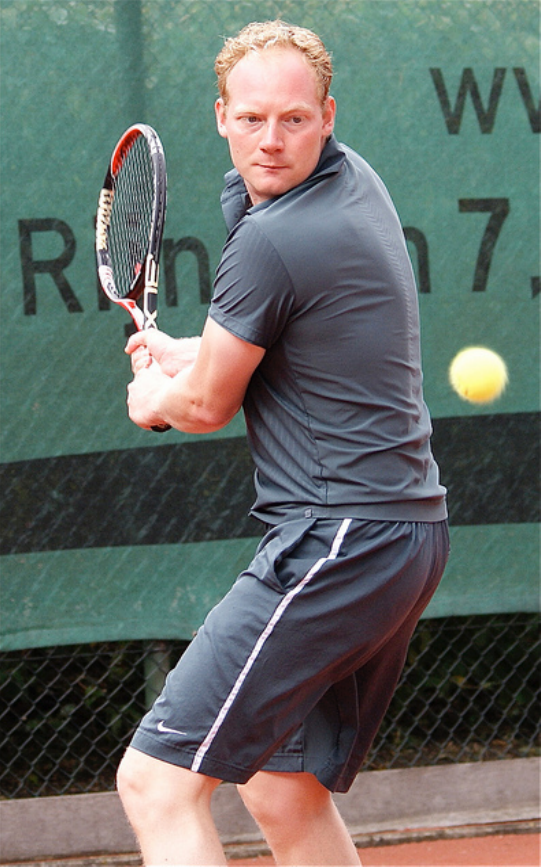}
\includegraphics[width=\widvis\linewidth]{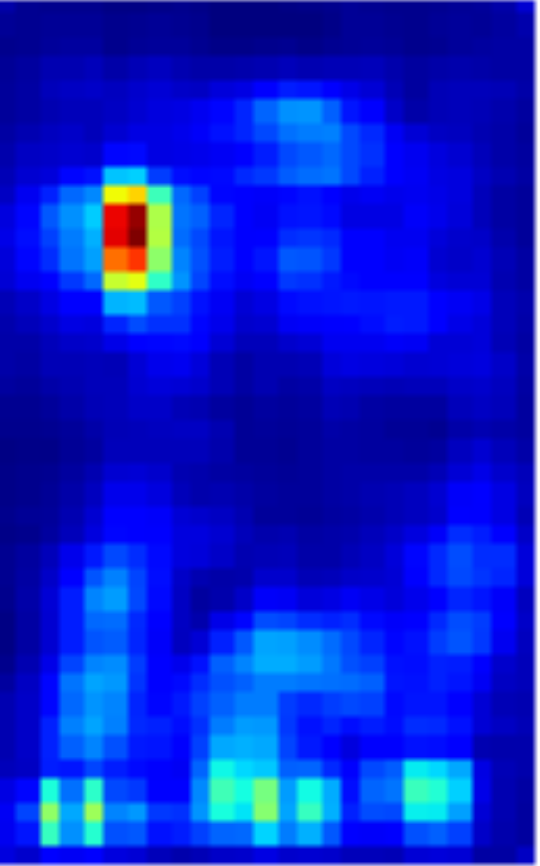}
\includegraphics[width=\widvis\linewidth]{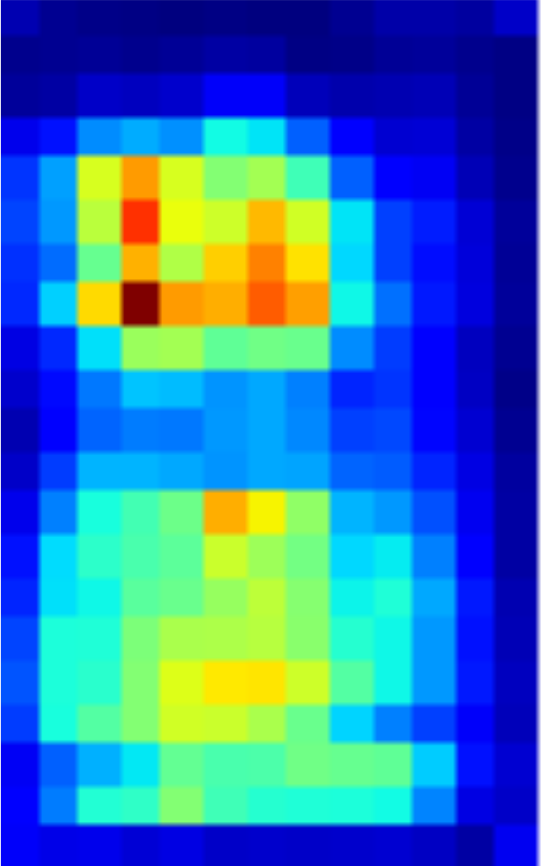}
\includegraphics[width=\widvis\linewidth]{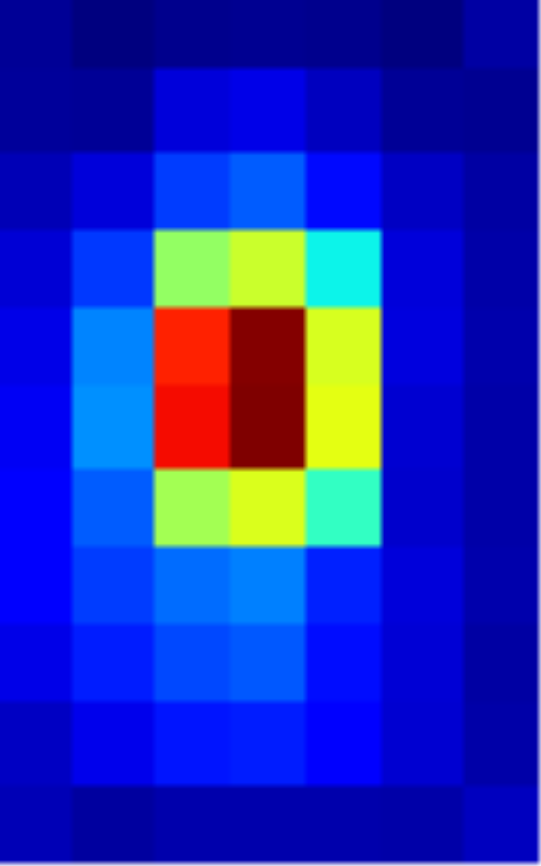} %
\includegraphics[width=\widvis\linewidth]{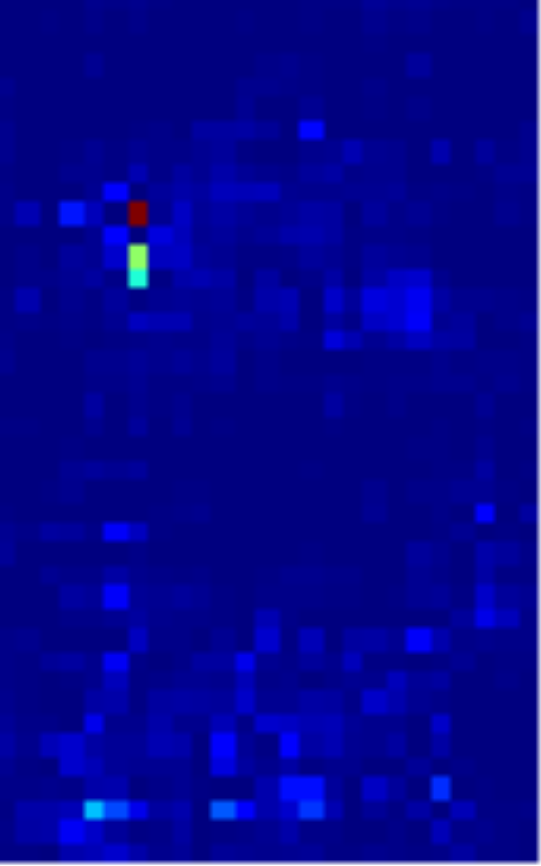}
\includegraphics[width=\widvis\linewidth]{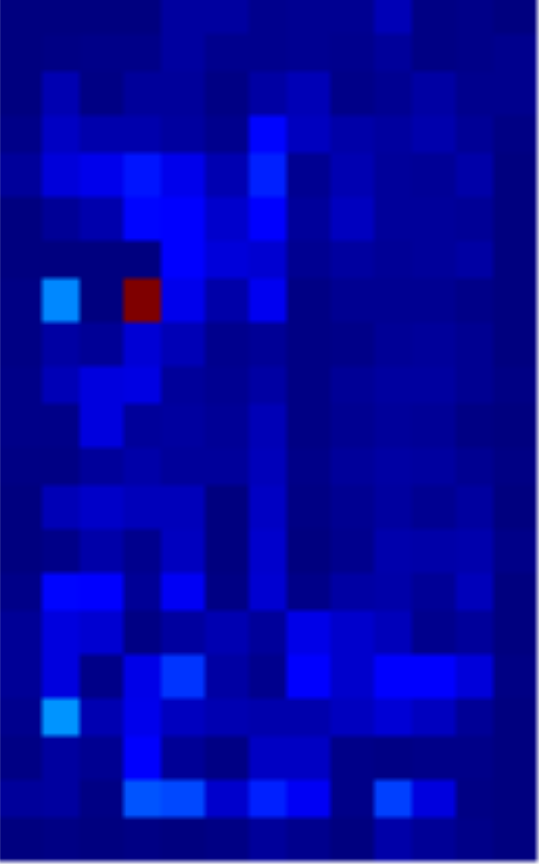}
\includegraphics[width=\widvis\linewidth,height=0.22165\linewidth]{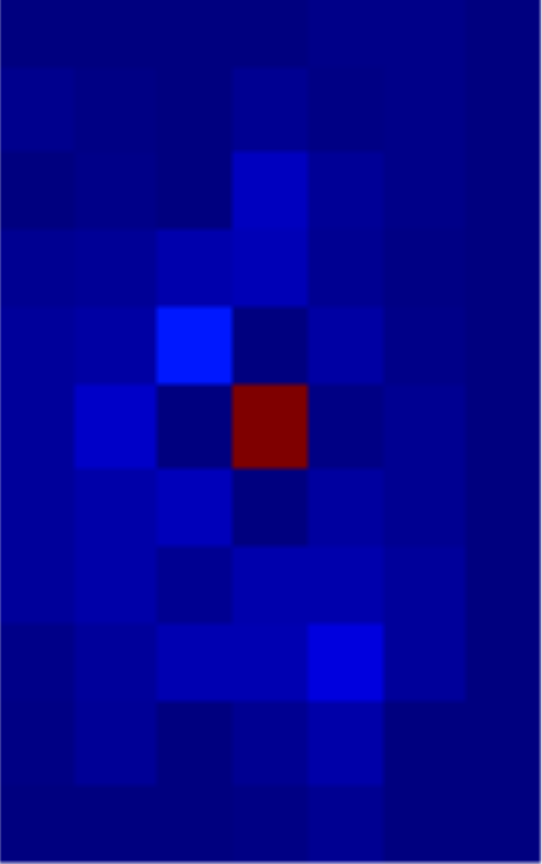}
\caption{Visualization of classification scores for different methods. 1st: input image.
2nd-4th depict the classification score heatmaps of the FCOS baseline;
5th-7th depict the %
score 
heatmaps of our \Ours. 
The three maps correspond to FPN levels of 
P5, P6, and P7.
Our \Ours\ trained with stop-gradient can significantly suppress duplicated  predictions.}
\label{fig:vis_scores}
\end{figure*}

\subsubsection{Effect of %
 Stop Gradient 
}

In this section, we analyze the effect of the asymmetric optimization scheme, \ie, stop the gradient relevant to the attached PSS head passing to the original FCOS network parameters. 
As depicted in Fig.~\ref{fig:model}, 
we use the features of the regression branch to train the PSS head. 
The vanilla solution is not to have the {\tt stop-grad} operation and update all the network parameters as usual.

Specifically, our \OursATSS\  method without stop-gradient %
achieves 41.6\%   mAP  (w/o NMS), which  outperforms DeFCN by only 0.1 points
and there is still a performance gap of 1.2 points
compared to the \RetinaNet\  baseline (42.8\%).
By applying the stop-gradient operation, \OursATSS\  achieves 42.6\%, almost the same as 
the baseline's 42.8\% mAP. 
We make %
a similar observation for the  \Ours\   method and it is worth noting that the end-to-end prediction result (w/o NMS) with stop-gradient achieves 42.3\%, which even improves the performance by 0.3\%  against the NMS-based FCOS baseline.

\begin{figure}[t!]
\centering 
\includegraphics[width=0.905\linewidth]{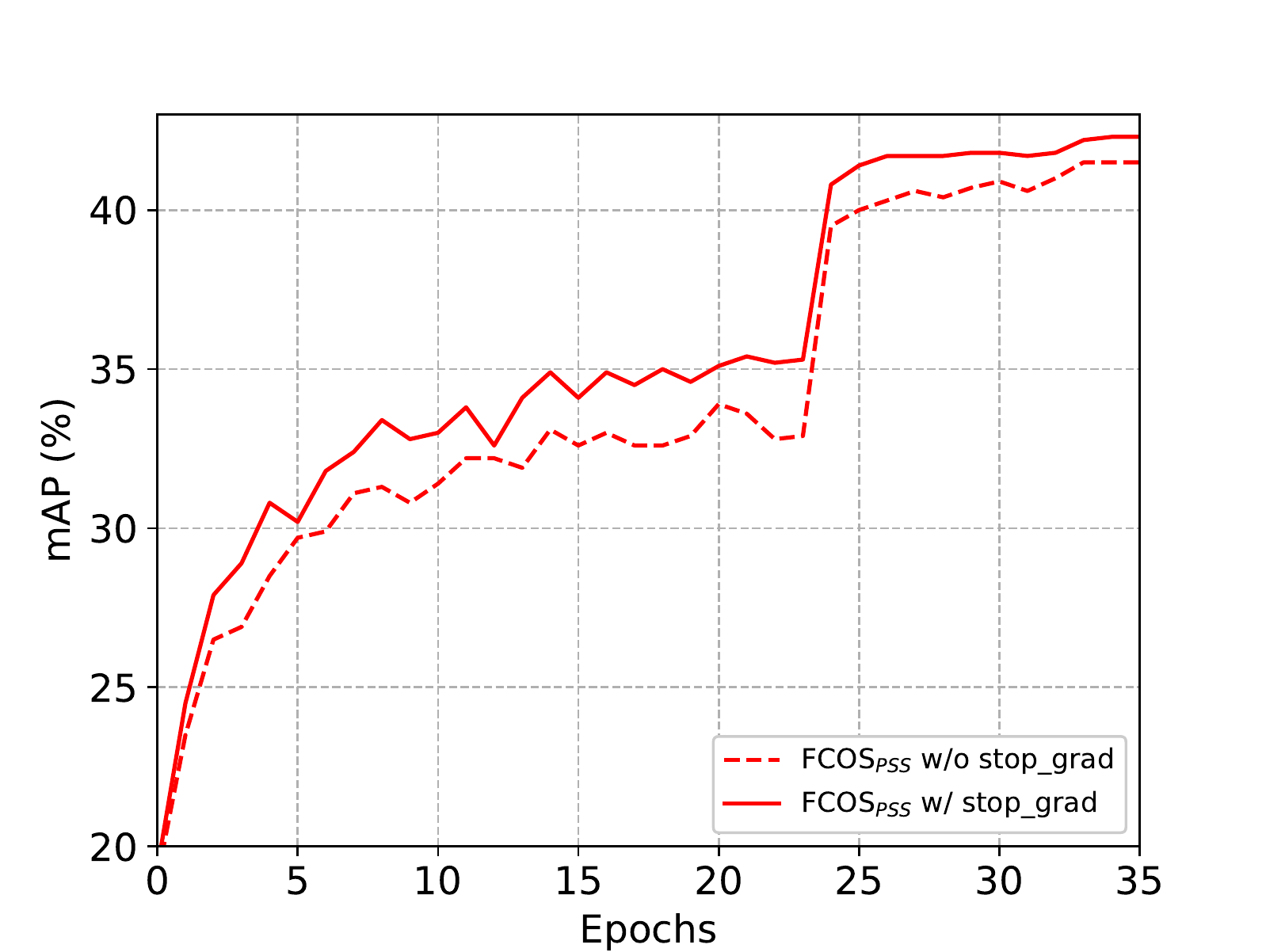}
\includegraphics[width=0.905\linewidth]{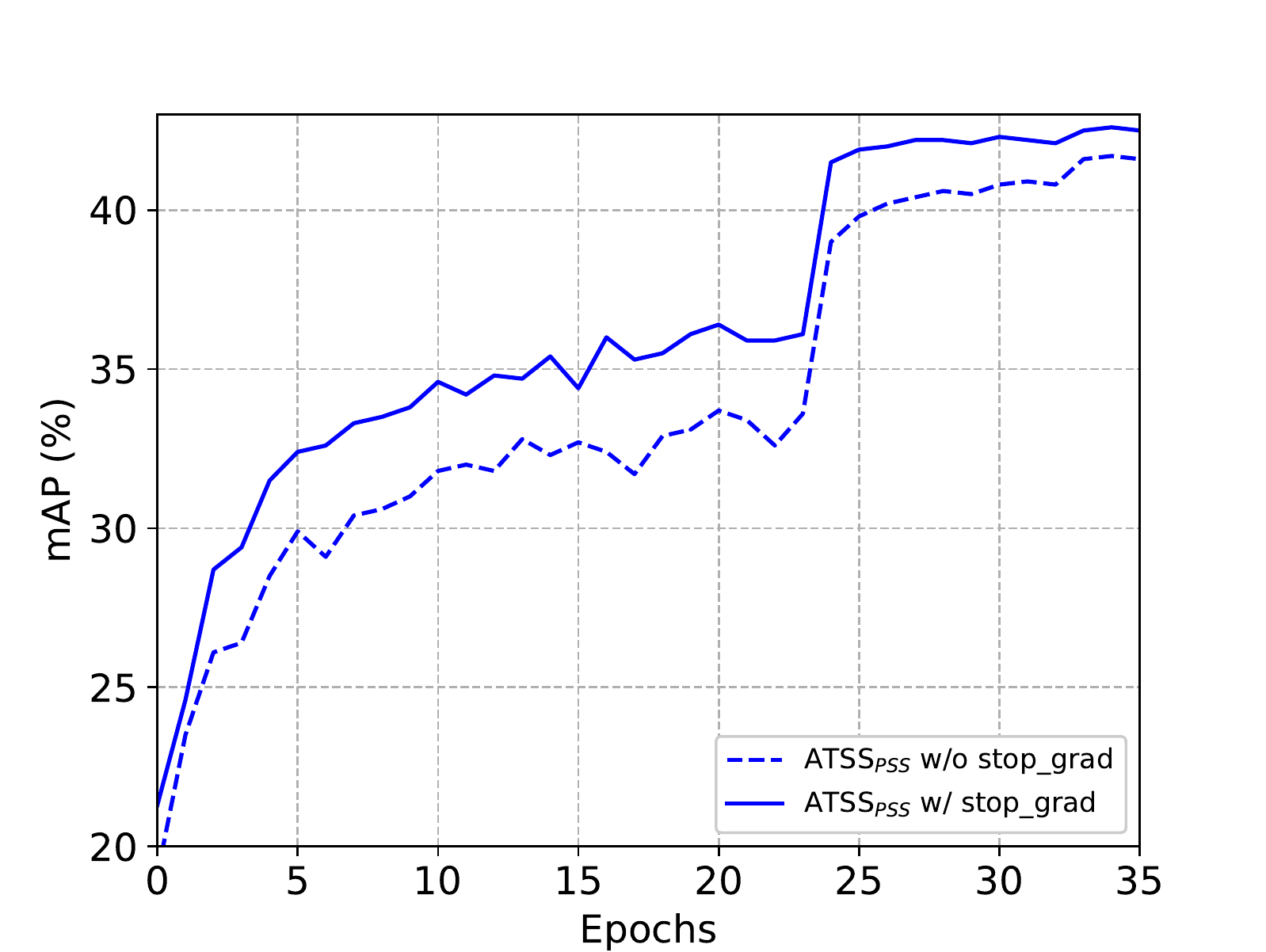}
\caption{Comparison of mAP for our \Ours\ (top) and  %
\OursATSS 
(bottom) models trained with and without stop-gradient w.r.t.\  training epochs, showing that
stop-gradient  helps training a better model.
}
\label{fig:stopgrad}
\end{figure}

To further verify whether stop-gradient can %
better train
our end-to-end detectors, we %
show the convergence curves 
of our \Ours\ and \OursATSS models trained with and without the stop-gradient operation on
the 
COCO val.\ set. As shown in Fig.~\ref{fig:stopgrad}, we can %
observe that stop-gradient %
indeed 
help train a better model and is able to consistently improve the performance during the training process.

We present the visualization of the predicted classification scores from the FCOS baseline and our \Ours\  methods in Fig.~\ref{fig:vis_scores}. The 2nd-4th depict the classification score heatmaps of the FCOS baseline and the 5th-7th depict the score heatmaps of our \Ours. The three score maps correspond to different FPN levels (P5, P6, P7).
It is clear 
that compared with FCOS baseline, our \Ours\ trained with stop-gradient is capable of significantly suppressing duplicate predictions.

\begin{table}[t!]
\begin{center}
\small 
    \begin{tabular}{r| r |c}
    \toprule
    Model          & Training Mode & \begin{tabular}[c]{@{}l@{}}mAP (w/o NMS)\end{tabular} \\ \midrule
    \OursATSS & end-to-end    & 42.6                                                    \\ \hline
    \OursATSS & two-step      & 42.3                                \\ \bottomrule
    \end{tabular}
\end{center}
\caption{Comparison of detection accuracy of our \OursATSS model when trained end-to-end vs.\ 
with two-step training.}
\label{tbl:train_modes}
\end{table}

\begin{table*}[h!]
\centering
\footnotesize 
    \begin{tabular}{ r  | l  |c|c|c|c}
    \toprule
    \multirow{3}{*}{Model}    & \multirow{3}{*}{Position}       & \multicolumn{2}{c|}{mAP (\%)}  & \multicolumn{2}{c}{mAR (\%)}      \\ \cline{3-6} 
                              &                                 & \begin{tabular}[c]{@{}c@{}}end-to-end pred.\\ (w/o NMS)\end{tabular} & \begin{tabular}[c]{@{}c@{}}one-to-many pred.\\ (w/ NMS)\end{tabular}  & \begin{tabular}[c]{@{}c@{}}end-to-end pred.\\ (w/o NMS)\end{tabular} & \begin{tabular}[c]{@{}c@{}}one-to-many pred.\\ (w/ NMS)\end{tabular}\\ \midrule
    \multirow{2}{*}{  \Ours\  } & PSS\ on regress.\ branch \             &    42.3       &     42.2    & 61.6  &    59.6      \\ 
                                & PSS\ on classif.\  branch \            &    41.5       &     42.2    & 60.9  &    59.4      \\ \midrule
    \multirow{2}{*}{  \OursATSS\   } &  PSS\ on regress.\                &    42.6       &     42.3    & 62.1  &    59.7      \\  
                                   & PSS\ on clssif.\                    &    41.8       &     42.3    & 61.4  &    60.0      \\ \bottomrule
    \end{tabular}
\caption{Comparison of detection accuracy for different methods by attaching the PSS head either to the classification branch or regression branch.
As we can see that the PSS head attached to the regression branch works slightly better. Therefore, in all other experiments we have attached the PSS head to the regression branch.
}
\label{tbl:ablat_cls_or_reg}
\end{table*}

\subsubsection{Stop Gradient vs.\ Two-step Training}
\label{subsec: stopvstwostep}
Stop-gradient allows the PSS module to be optimized 
\textit{asymmetrically} compared to the rest of the network; 
and may benefit from the feature representation learning of the original FCOS/ATSS model. If we train in two steps, \ie, first train the FCOS/ATSS model to convergence, and then train the PSS module alone by freezing the FCOS/ATSS model, 
    would this two-step strategy work well?  
    As discussed in \S\ref{subsec:stopgrad}, this two-step optimization corresponds to 
    optimization of sub-problems of \eqref{eq:alt1} and 
    \eqref{eq:alt2} to convergence for only 
    one alternation.  

We train such a model by training a baseline model in step one for 36 epochs, and 
then train the PSS head in step two for another 12 epochs. 
As shown in Table~\ref{tbl:train_modes}, the two-step training of %
the \OursATSS model %
achieves 42.3\% mAP (vs.\ 42.6\% for end-to-end training).

\begin{figure}[t!]
\centering 
\includegraphics[width=0.905\linewidth]{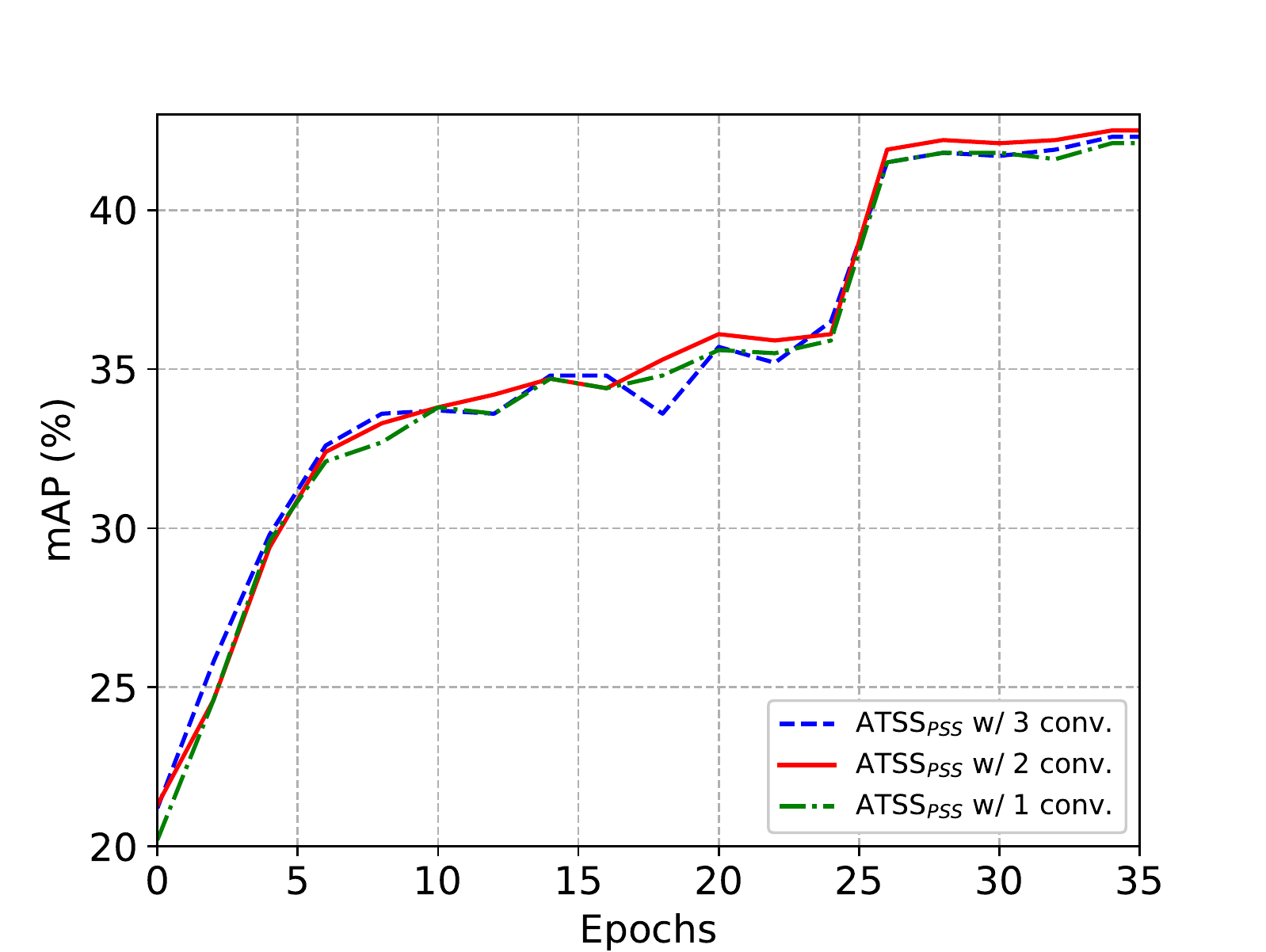}
\caption{Comparison of mAP for our \OursATSS method trained with different conv.\ layers of the 
PSS head w.r.t.\ training epochs.}
\label{fig:convlayers}
\end{figure}

\subsubsection{Attaching PSS to Regression  vs.\  Classification Branch}

In this section, we analyze the effect of attaching the PSS head either to the classification branch or the regression branch. As shown in Table~\ref{tbl:ablat_cls_or_reg}, compared with attaching to the regression branch, moving the PSS head to the classification branch results in performance degradation of end-to-end prediction, while the performance of one-to-many prediction remains unchanged. 
We %
thus
conclude that it is more suitable to optimize the PSS head with features of the regression branch. Unless otherwise specified, we attach the PSS head to the regression branch in all other experiments. Recall that in the original FCOS, it was also observed that attaching the centre-ness head to regression works better than attaching to the classification branch.

\subsubsection{How Many Conv.\ Layers for the PSS Head}
Since we expect to learn a binary classifier for one-to-one prediction, 
the capacity of the head may play an important role.
We use standard conv.\ layers to facilitate the learning of the binary mask. To verify how many conv.\ layers %
work well, 
we perform ablation study w.r.t.\  the number of conv.\ layers based on \Ours\ and \OursATSS. 
As shown in Table~\ref{tbl:ablation_convs}, %
the PSS head with two conv.\ layers achieves satisfactory accuracy. 
It is interesting to see that increasing the PSS head to three conv.\ layers 
slightly deteriorates the detection performance, which may be due to overfitting.
We plot the convergence curves in  Fig.~\ref{fig:convlayers}, showing that at the beginning, 
PSS with three layers obtains better mAP but later becomes slightly worse than that of two layers. Although carefully tuning the hyper-parameters may 
result in slightly different results, we may conclude here that two conv.\ layers for the PSS head already work well.

\begin{table*}[h!]
\centering 
\footnotesize  
        \begin{tabular}{ r |c|cc|cc}
        \toprule
        \multirow{2}{*}{Model} & \multirow{2}{*}{$\#$ Conv.\ layers } & \multicolumn{2}{c|}{mAP (\%)}                  & \multicolumn{2}{c}{mAR (\%)} \\
                               &                         & w/o NMS     & w/ NMS     & w/o NMS     & w/ NMS  \\
        \midrule
        \multirow{3}{*}{  \Ours\  }          &        1      &  41.6         & 41.9    &  61.4 & 59.5 \\
                                           &        2      & \textbf{42.3} & 42.2    &  61.6 & 59.6 \\
                                           &        3      &  41.9         & 41.7    &  61.7 & 59.2 \\
        \midrule
        \multirow{3}{*}{  \OursATSS  }          &        1      &   42.1        &  42.0    & 61.7  &  59.3 \\
                                           &        2      & \textbf{42.6} & 42.3    & 62.1  & 59.7  \\
                                           &        3      &   42.3        &  41.7    &  61.9  & 59.3  \\
        \bottomrule
        \end{tabular}
\caption{The effect of the number of conv.\ layers for the PSS head. Each conv.\ layer has 256 channels with stride 1. All models are based on ResNet-50 with FPN; and the `3\x' training schedule. In general, two conv.\ layers show satisfactory  accuracy.
}
\label{tbl:ablation_convs}
\end{table*}

\begin{table}[t!]
\begin{center}
\footnotesize 
    \begin{tabular}{ r  | c|c|c}
    \toprule
    \multirow{3}{*}{Model}    & \multirow{3}{*}{Center-ness} & \multicolumn{2}{c}{mAP (\%)}  \\ \cline{3-4} 
                              &                                  & \begin{tabular}[c]{@{}c@{}}end-to-end pred.\\ (w/o NMS)\end{tabular} & \begin{tabular}[c]{@{}c@{}}
                              one-to-many
                              pred.\\ (w/ NMS)\end{tabular} \\ \midrule
    \multirow{2}{*}{  \Ours\  } &                                  &     41.7      & 41.4     \\ 
                              & \checkmark                         &  42.3     &  42.2   \\ \midrule
    \multirow{2}{*}{  \OursATSS  } &                               &    42.1   &    41.7  \\  
                              & \checkmark                         &   42.6    &   42.3   \\ \bottomrule
    \end{tabular}
\end{center}
\caption{Comparison of detection accuracy for different models trained with and without the center-ness %
head 
respectively.
}
\label{tbl:ablat_centerness}
\end{table}

\subsubsection{Effect of the Center-ness Branch}

To evaluate the effect of the center-ness branch, we perform ablation study on our end-to-end detectors \Ours\ and \OursATSS trained with and without center-ness branch, respectively. 

{Compared with the positive samples far from the center of the instance, it %
make sense 
for the positive samples close to the center of the instance to %
have 
a larger loss weight. Therefore, although we remove the centerness branch, we still retain the centerness as the loss weight of GIoU loss during the training process.}
We report the results in Table \ref{tbl:ablat_centerness}. 
For \Ours,  center-ness improves the mAP by 0.6 points. This is consistent with the observation for the FCOS baseline (see \S3.1.3 of \cite{FCOSPAMI}).

For \OursATSS, center-ness can  improve by 0.5 points. This is expected as a better 
positive example sampling strategy slightly diminishes the effectiveness the centerness. 

\begin{table}[t!]
\begin{center}
\footnotesize 
\begin{tabular}{ r  | c|c|c}
\toprule
\multirow{3}{*}{Model}    & \multirow{3}{*}{$\lambda_1$} & \multicolumn{2}{c}{mAP (\%)}  \\ \cline{3-4} 
                          &                                  & \begin{tabular}[c]{@{}c@{}}
                          end-to-end
                          pred.\\ (w/o NMS)\end{tabular} & \begin{tabular}[c]{@{}c@{}}
                          one-to-many
                          pred.\\ (w/ NMS)\end{tabular} \\ \midrule
\multirow{3}{*}{  \Ours\  } &     0.5           &   41.7    &   41.9  \\ 
                            &     1.0           &   \textbf{42.3}    &  42.2   \\
                            &     2.0           &    41.7   &  41.5   \\ \bottomrule
\end{tabular}
\end{center}
\caption{Comparison of detection accuracy by varying the loss weight of
the PSS term as in Equ.~\eqref{eq:overall_loss}  
on the COCO val.\  set.
}
\label{tbl:ablat_lossweight}
\end{table}

\subsubsection{Loss Weight of PSS Loss $\Loss_{pss}$}

In this section, we analyze the sensitivity of the loss weight $\lambda_1$ of the PSS loss $\Loss_{pss}$ in Equ.~\eqref{eq:overall_loss}. Specifically, we evaluate our \Ours\ method with different values %
for 
$\lambda_1 \in \{0.5, 1.0, 2.0\}$ and report the results in Table~\ref{tbl:ablat_lossweight}. It appears that $ \lambda_1 = 1 $ already works well.

\begin{table}[t!]
\centering 
\small 
    \begin{tabular}{l|c|c|c}
    \toprule
    Method               & $\alpha$ & mAP (\%)  & mAR (\%) \\ 
    \midrule
    \multirow{4}{*}{Add} & 0.2                   & 41.6 & 61.3 \\ 
                         & 0.4                   & 41.7 & 61.4 \\ 
                         & 0.6                   & 41.7 & 60.9 \\ 
                         & 0.8                   & 41.3 & 60.6 \\ 
    \midrule
    \multirow{4}{*}{Mul.\ } & 0.2                   & 41.6 & 60.4 \\ 
                         & 0.4                   & 41.8 & 61.1 \\ 
                         & 0.6                   & 41.8 & 61.0 \\ 
                         & 0.8                   & \textbf{42.3} & \textbf{61.6} \\ \bottomrule
    \end{tabular}
\caption{Comparison of different matching score functions of our \Ours model on the COCO val.\ set.}
\label{tbl:ablat_matchfunc}
\end{table}

\subsubsection{Matching Score Function}

To explore the best formulation of matching score functions of our \Ours model, we borrow the `Add' function form of~\cite{wang2020end_defcn} and replace the matching score function in Equ.~\eqref{eq:quality} by
$(1-\alpha) \cdot \hat{P}_{i}(c_j) + \alpha \cdot {\rm IoU}(\hat{b}_{i}, b_j)$.
As shown in Table~\ref{tbl:ablat_matchfunc}, compared with `Add', the `Mul.' function 
works slightly better, 
achieving 
the best detection performance of 42.3\%  mAP with $\alpha=0.8$.
Therefore, we use `Mul.' function with $\alpha=0.8$ in all other experiments.

\subsubsection{Effect of the Ranking Loss}

To demonstrate the effectiveness of the ranking loss, we performs experiments on our \Ours\ and \OursATSS detectors trained with and without ranking loss respectively.
The results are reported in Table~\ref{tbl:rankingloss}, without introducing much training complexity, the ranking loss can further improves the final detection performance by 0.2\app 0.3 points in mAP. 

\begin{table}[h!]
\centering 
\small
    \begin{tabular}{ r |c| c }
    \toprule
    Model                     & Ranking loss & mAP (\%)  \\ 
    \midrule
    \multirow{2}{*}{  \Ours\  } &             & 42.0 \\ 
                              &    %
                                   \checkmark
                                             & 42.3 \\ 
    \midrule
    \multirow{2}{*}{  %
                      \OursATSS } &             & 42.4 \\ 
                              &     %
                                    \checkmark
                                       & 42.6 \\ 
    \bottomrule
    \end{tabular}
\caption{Comparison of mAP for our \Ours\ and \OursATSS models trained with and without the ranking loss, indicating that our ranking loss can further improve the detection performance.}
\label{tbl:rankingloss}
\end{table}

\begin{figure*}
\centering 
{ 
\includegraphics[trim=0.5cm .75cm 0 0,clip,width=0.905\linewidth]{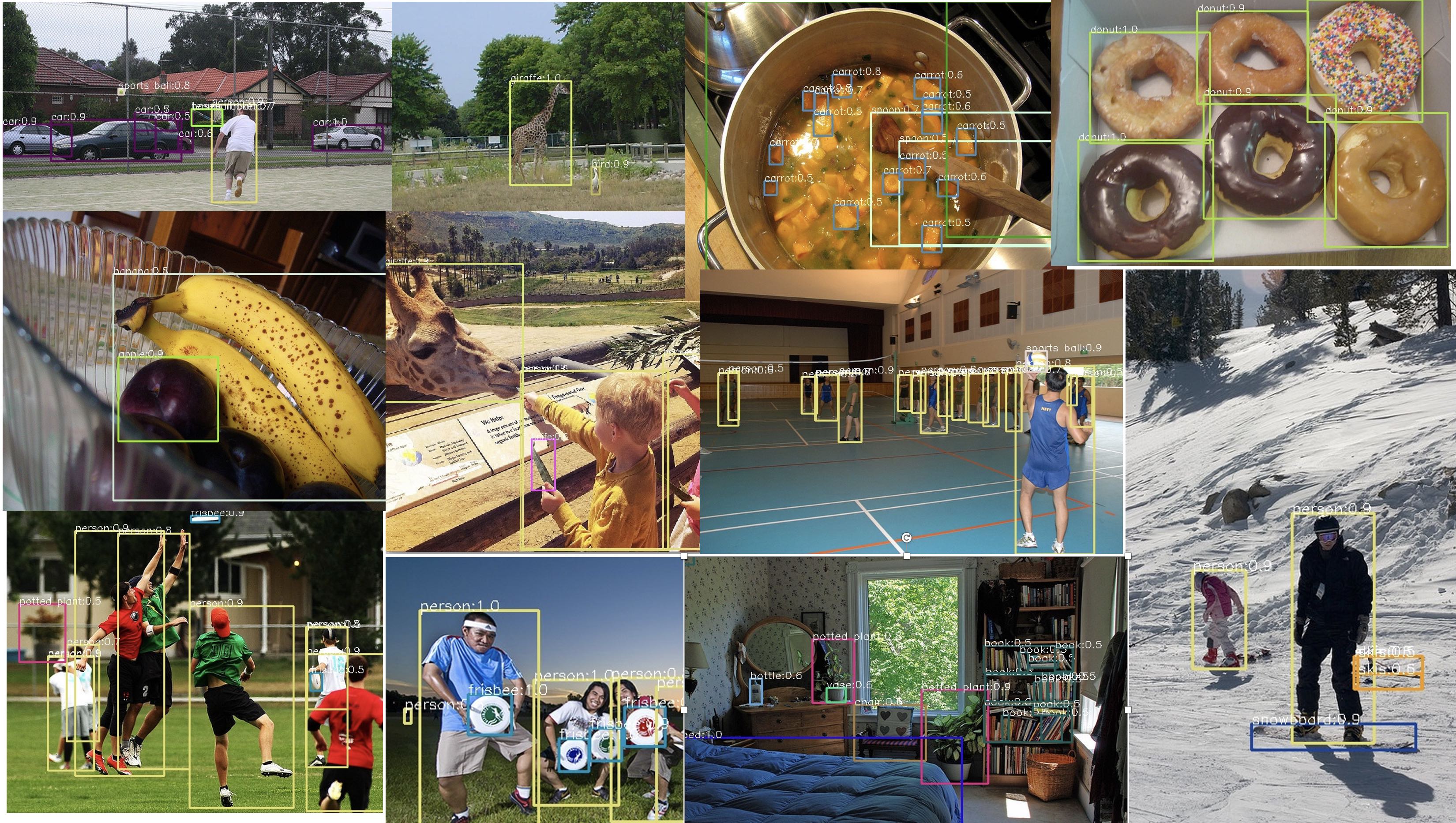}
}
\caption{Visalization of some detection results from the COCO dataset.
Results are obtained using the \OursATSS with the R2N-101-DCN backbone model, achieving 48.5\% mAP on the COCO val.\ set.
}
\label{fig:vis_overall}
\end{figure*}

\subsection{Comparison with State-of-the-art}

We compare \Ours/\OursATSS  with other state-of-the-art object detectors on the COCO benchmark. For these
experiments, we make use
of multi-scale training. %
In particular, during training, the
shorter side of the input image is sampled from  $ [480, 800] $
with a step of 32 pixels.

From Table \ref{tab:main}, we make the following conclusions.
\begin{itemize}
\itemsep -0.15cm
    \item 
    Compared to the standard baseline FCOS, the proposed \Ours{} 
    even achieves slightly detection accuracy, with negligible computation overhead.

    The detection mAP difference between the baseline ATSS and the proposed \OursATSS 
    is less than 0.2 points. 
    \item 
    Our methods outperform the recent end-to-end NMS-free detector DeFCN.
    
    \item 
    Our methods show improved performance with backbones of large capacity.
    In particular, we have trained models of ResNeXt-32\x4d-101 with deformable convolutions,
    and Res2Net-101 with deformable convolutions. 
    Our models achieve 47.5\% and 48.5\% mAP respectively, which are among the state-of-the-art.

\end{itemize}
 We show some qualitative results in Fig.~\ref{fig:vis_overall}.

\begin{table}[h!]
\centering 
\footnotesize 
    \begin{tabular}{r|c|c|c}
    \toprule
    Model                   & Sched. & Box mAP (\%) & %
    Mask mAP (\%) \\
    \midrule
    CondInst~\cite{tian2020conditional}                & 1$\times$     & 38.9     & 34.1      \\
    CondInst~\cite{tian2020conditional}                & 3$\times$     & 42.1     & 37.0      \\
    \midrule
    \textbf{\OursCondInst} (Ours)    & 1$\times$     & 39.4     & 34.4      \\
    \textbf{\OursCondInst} (Ours)    & 3$\times$     & 42.4     & 36.8      \\
    \bottomrule
    \end{tabular}
\caption{Comparison of mAP for the CondInst baseline and our \OursCondInst\ models on the COCO val.\  set.
The backbone is R50.}
\label{tbl:condinstpss}
\end{table}

\subsection{PSS for Instance Segmentation}
\label{exp:inst}

{
Instance segmentation is a fundamental yet challenging task in computer vision and has received
much 
attention %
in
the community~\cite{he2017mask,tian2020conditional}.
In this section we demonstrate that our PSS head can benefit the NMS-based Instance segmentation methods. Recently, CondInst~\cite{tian2020conditional} eliminates ROI operations by %
employing 
dynamic instance-aware networks, which achieves improved instance segmentation performance. However, CondInst still needs box-based NMS to achieve accurate and fast performance during inference. Similar to \Ours, we only introduce our PSS head to the original CondInst network, denoted as \OursCondInst. We
carry out 
experiments on the COCO dataset and report the results in Table~\ref{tbl:condinstpss}. The results indicate that by eliminating heuristic NMS, we can simplify the instance segmentation method while still 
achieving competitive performance.

}

\section{Conclusion}

In this work, we have proposed a very simple modification to the original FCOS detector (and its improved version of ATSS) for eliminating the heuristic NMS post-processing. 
We achieve that by attaching a compact positive sample selector head to the bounding box regression branch of the FCOS network, which consists of only two conv.\ layers. 
Thus, once trained, the new detector can be deployed easily without any heuristic processing involved. The proposed detector is made simpler and end-to-end trainable. 
We also show that the same idea can be applied to instance segmentation for removing NMS.
We expect to 
see that the proposed design in this work may benefit many other instance-level recognition tasks beyond bounding-box object detection.

{\small
\bibliographystyle{ieee_fullname}
\bibliography{main}
}

\end{document}